\documentclass[10pt,onecolumn,letterpaper]{article}

\usepackage{datadog}

\usepackage{amssymb}
\usepackage{booktabs}
\usepackage{multirow}
\usepackage{multicol}

\begin{document}
\title{Toto: Time Series Optimized \\ Transformer for Observability \\ [10pt] \large\textit{Technical Report}\\[10pt]}

\author{
Ben Cohen \and Emaad Khwaja \and Kan Wang \and Charles Masson \AND Elise Ram\'e 
\and Youssef Doubli \and Othmane Abou-Amal \AND \hphantom{me} 
{\tt\small \{ben.cohen, emaad, kan.wang, charles.masson, elise.rame, youssef.doubli, othmane\}@\href{https://datadoghq.com}{datadoghq.com}}
}

\maketitle

\begin{abstract}
This technical report describes the Time Series Optimized Transformer for Observability (Toto), a new state-of-the-art foundation model for time series forecasting developed by Datadog. In addition to advancing the state of the art on generalized time series benchmarks in domains such as electricity and weather, this model is the first general-purpose time series forecasting foundation model to be specifically tuned for observability metrics.

Toto was trained on a dataset of one trillion time series data points – the largest among all currently published time series foundation models. Alongside publicly available time series datasets, 75\% of the data used to train Toto consists of  fully anonymous numerical metric data points from the Datadog platform.	

In our experiments, Toto outperforms existing time series foundation models on observability data. It does this while also excelling at general-purpose forecasting tasks, achieving state-of-the-art zero-shot performance on multiple open benchmark datasets.

\hfill \break
In this report, we detail the following key contributions:
\begin{itemize}
    \item \textbf{Proportional factorized space-time attention:} We introduce an advanced attention mechanism that allows for efficient grouping of multivariate time series features, reducing computational overhead while maintaining high accuracy.
    \item \textbf{Student-T mixture model head:} This novel use of a probabilistic model that robustly generalizes Gaussian mixture models enables Toto to more accurately capture the complex dynamics of time series data and provides superior performance over traditional approaches.
    \item \textbf{Domain-specific training data:} In addition to general multi-domain time series data, Toto is specifically pre-trained on a large-scale dataset of Datadog observability metrics, encompassing unique characteristics not present in open-source datasets. This targeted training ensures enhanced performance in observability metric forecasting.
\end{itemize}

\begin{figure*}[!htbp]
  \centering
    \includegraphics[width=.85\linewidth]{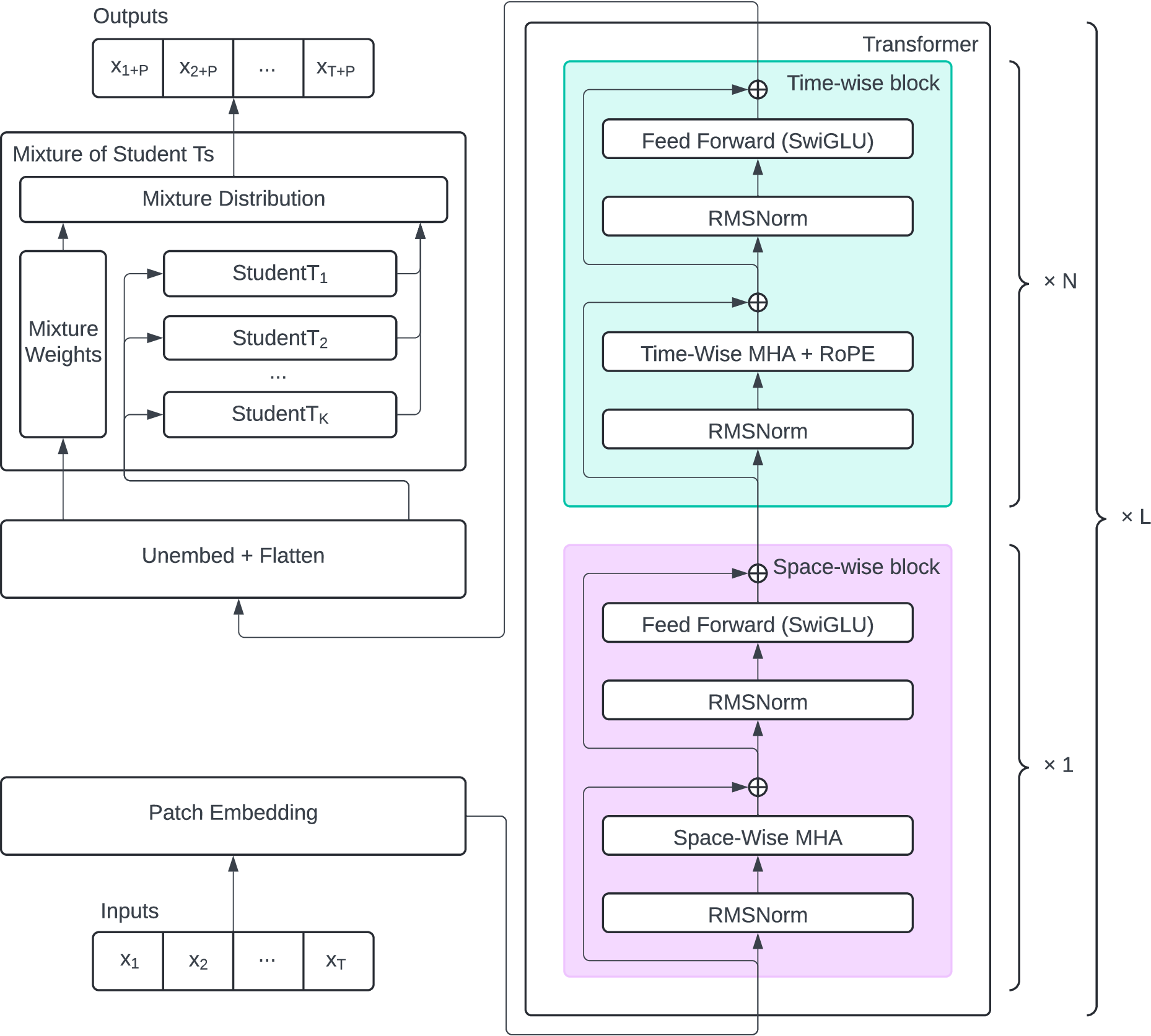}
   \caption{
    Toto architecture diagram. Input time series of \( T \) steps (univariate example used for simplicity here) are first embedded using the patch embedding layer. They then pass through the transformer stack, which contains \( L \) identical segments. Each segment of the transformer consists of one space-wise transformer block followed by \( N \) time-wise blocks. The flattened transformer outputs are projected to form the parameters of the Student-T mixture model (SMM) head. The final outputs are the forecasts for the input series, shifted \( P \) steps (the patch width) into the future.
    }
   \label{fig:architecture}
\end{figure*}

\end{abstract}

\begin{multicols}{2}
\section{Background} \label{background}
\begin{figure*}[!ht]
  \centering
   \includegraphics[width=1\linewidth]{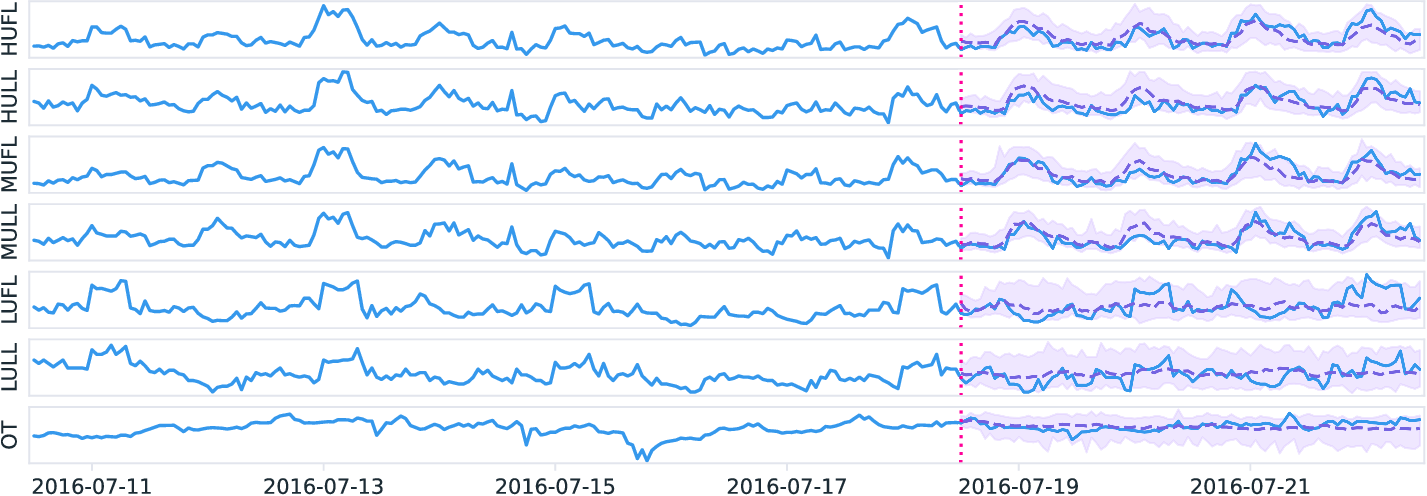}

   \caption{Example of Toto\textquotesingle s 96-step zero-shot forecasts on the ETTh1 dataset, showing multivariate probabilistic predictions. Solid lines represent ground truth, dashed lines represent median point forecasts, and shaded regions represent 95\% prediction intervals.}
   \label{fig:forecast}
\end{figure*}

We present Toto, a groundbreaking time series forecasting foundation model developed by Datadog. Toto is specifically designed to handle the complexities of observability data, leveraging a state-of-the-art transformer architecture to deliver unparalleled accuracy and performance. Toto is trained on a massive dataset of diverse time series data, enabling it to excel in zero-shot predictions. This model is tailored to meet the demanding requirements of real-time analysis as well as compute and memory-efficient scalability to very large data volumes, providing robust solutions for high-frequency and high-dimensional data commonly encountered in observability metrics.

\subsection{Observability data}
The Datadog observability platform collects a vast array of metrics across multiple subdomains, crucial for monitoring and optimizing modern infrastructure and applications. These metrics include infrastructure data such as memory usage, CPU load, disk I/O, and network throughput, as well as application performance indicators like hit counts, error rates, and latency \cite{Datadog_Observability}. Additionally, Datadog integrates specific metrics from numerous SaaS products, cloud services, open-source frameworks, and other third-party tools. The platform allows users to apply various time series models to proactively alert on anomalous behavior, leading to a reduction in time to detection (TTD) and time to resolution (TTR) of production incidents \cite{Datadog_Infrastructure}. 

The complexity and diversity of these metrics present significant challenges for time series forecasting. Observability data often requires high time resolution, down to seconds or minutes, and is typically sparse with many zero-inflated metrics. Moreover, these metrics can display extreme dynamic ranges and right-skewed distributions. The dynamic and nonstationary nature of the systems being monitored further complicates the forecasting task, necessitating advanced models that can adapt and perform under these conditions.

\subsection{Traditional models}
Historically, time series forecasting has relied on classical models such as ARIMA, exponential smoothing, and basic machine learning techniques \cite{Hyndman2021}. While foundational, these models necessitate individual training for each metric, presenting several limitations \cite{Fildes1998}. The need to develop and maintain separate models for each metric impedes scalability, especially given the extensive range of metrics in observability data. Moreover, these models often fail to generalize across different types of metrics, leading to suboptimal performance on diverse datasets \cite{Stevenson2007, Christodoulos2010}. Continuous retraining and tuning to adapt to evolving data patterns further increase the operational burden.  This scaling limitation has hindered the adoption of deep learning–based methods for time series analysis, even as they show promise in terms of accuracy \cite{Salinas2020}.

\subsection{Foundation models}
Large neural network-based generative models, often referred to as ``foundation models,'' have revolutionized time series forecasting by enabling accurate predictions on new data not seen during training, known as zero-shot prediction \cite{Brophy2023}. This capability significantly reduces the need for constant retraining on each specific metric, thus saving considerable time and computational resources. Their architecture supports the parallel processing of vast data volumes, facilitating timely insights essential for maintaining system performance and reliability \cite{Jia2018, Xu2021}.

Through pretraining on diverse datasets, generative models exhibit strong generalization across various types of time series data. This enhances their robustness and versatility, making them suitable for a wide range of applications. Zero-shot predictions are particularly attractive in the observability domain, where the limitations of traditional methods are felt very acutely. The most common use cases for time series models within an observability platform like Datadog include automated anomaly detection and predictive alerting. It is challenging to scale classical forecasting methods to handle cloud-based applications that can be composed of many ephemeral, dynamically scaling components such as containers, VMs, serverless functions, etc. These entities tend to be both high in cardinality and short-lived in time. This limits the practicality of traditional time series models in two ways:

\begin{itemize}

    \item First, the high cardinality and volume of data can make fitting individual models to each time series computationally expensive or even intractable. The ability to train a single model and perform inference across a wide range of domains has the potential to dramatically improve the efficiency, and thus the coverage, of an autonomous monitoring system.
    
    \item Second, ephemeral infrastructure elements often lack enough historical data to confidently fit a model. In practice, algorithmic alerting systems often require an adaptation period of days or weeks before they can usefully monitor a new metric. However, if the object being monitored is a container with a lifespan measured in minutes or hours, these classical models are unable to adapt quickly enough to be useful. Real-world systems thus often fall back to crude heuristics, such as threshold-based alerts, which rely on the domain knowledge of users. Zero-shot foundation models can enable accurate predictions with much less historical context, by aggregating and interpolating prior information learned from a massive and diverse dataset.

\end{itemize}

The integration of transformer-based models \cite{Vaswani2017} like Toto into observability data analysis thus promises significant improvements in forecasting accuracy and efficiency. These models offer a robust solution for managing diverse, high-frequency data and delivering zero-shot predictions. With their advanced capabilities, transformer-based models represent a significant leap forward in the field of observability and time series analysis \cite{Wu2021, Zhou2020, Nie2023}.

\subsection{Recent work}
The past several years have seen the rise of transformer-based models as powerful tools for time series forecasting. These models leverage multi-head self-attention mechanisms to capture long-range dependencies and intricate patterns in data.

To address the unique challenges of time series data, recent advancements have introduced various modifications to the attention mechanism. For example, Moirai \cite{Woo2024} uses ``any-variate'' attention to model dependencies across different series simultaneously. Factorized attention mechanisms \cite{zhang2023crossformer} have been developed to separately capture temporal and spatial (cross-series) interactions, enhancing the ability to understand complex interdependencies. Other models \cite{Liu2024,ilbert2024samformer} have used cross-channel attention in conjunction with feed-forward networks for mixing in the time dimension. Additionally, causal masking \cite{das2024a} and hierarchical encoding \cite{zhang2023crossformer} can improve the efficiency and accuracy of predictions in time series contexts.

These innovative transformer-based models have demonstrated state-of-the-art performance on benchmark datasets \cite{Nie2023}, frequently surpassing traditional models in both accuracy and robustness. Their capacity to process high-dimensional data efficiently \cite{Lin_2021_Survey} makes them ideal for applications involving numerous time series metrics with varying characteristics, such as observability.

Even more recently, a number of time series ``foundation models'' have been released \cite{das2024a, ansari2024chronoslearninglanguagetime, Woo2024, garza2023timegpt1, rasul2023lagllama, gruver2023large}. By pre-training on extensive, multi-domain datasets, these large models achieve impressive zero-shot prediction capabilities, significantly reducing the need for constant retraining. This paradigm is appealing for the observability context, where we constantly have new time series to process and frequent retraining is impractical.

\section{Problem statement} \label{problem_statement}
At Datadog, our time series data encompasses a variety of observability metrics from numerous subdomains. These metrics present several challenges for existing forecasting models:

\begin{itemize}
    \item \textbf{High time resolution:} Users often require data in increments of seconds or minutes, unlike many publicly-available time series datasets that are at hourly frequency or above.
    
\item \textbf{Sparsity:} Metrics such as error counts often track rare events, resulting in sparse and zero-inflated time series.

\item \textbf{Extreme right skew:} Latency measurements in distributed systems exhibit positive, heavy tailed distributions with extreme values at high percentiles.

\item \textbf{Dynamic, nonstationary systems:} The behavior of monitored systems changes frequently due to code deployments, infrastructure scaling, feature flag management, and other configuration changes, as well as external factors like seasonality and user-behavior-driven trends. Some time series, such as those monitoring fleet deployments, can also have a very low variance, exhibiting a piecewise-constant shape.

\item \textbf{High-cardinality multivariate data:} Monitoring large fleets of ephemeral cloud infrastructure such as virtual machines (VMs), containers, serverless functions, etc. leads to high cardinality data, with hundreds or thousands of individual time series variates, often with limited historical data for each group.

\item \textbf{Historical anomalies:} Historical data often contains outliers and anomalies caused by performance regressions or production incidents.
\end{itemize}

Foundation models pre-trained on other domains struggle to generalize effectively to observability data due to these characteristics. To overcome this, we developed Toto, a state-of-the-art foundation model that excels at observability forecasting while also achieving top performance on standard open benchmarks.

\section{Model architecture} \label{model_architecture}
Toto is a decoder-only forecasting model. This model employs many of the latest techniques from the literature, and introduces a novel method for adapting multi-head attention to multivariate time series data (\fref{fig:architecture}). 

\subsection{Transformer design}
Transformer models for time series forecasting have variously used encoder-decoder \cite{Zhou2020, Wu2021, ansari2024chronoslearninglanguagetime}, encoder-only \cite{Nie2023, Woo2024, Liu2024}, and decoder-only architectures \cite{rasul2023lagllama, das2024a}. For Toto, we employ a decoder-only architecture. Decoder architectures have been shown to scale well \cite{Radford2018ImprovingLU, Radford2019LanguageMA}, and allow for arbitrary prediction horizons. The causal next-patch prediction task also simplifies the pre-training process.

We use techniques from some of the latest large language model (LLM) architectures, including pre-normalization \cite{xiong2020on}, RMSNorm \cite{zhang-sennrich-neurips19}, and SwiGLU feed-forward layers \cite{shazeer2020gluvariantsimprovetransformer}.

\subsection{Input embedding}
Time series transformers in the literature have used various approaches for creating input embeddings. We use non-overlapping patch projections (\fref{fig:patch_embedding}), first introduced for Vision Transformers \cite{DBLP:conf/iclr/CordonnierLJ20, dosovitskiy2021an} and popularized in the time series context by PatchTST \cite{Nie2023}. Toto was trained using a fixed patch size of 32. 

\begin{minipage}{\linewidth}
    \centering
    \includegraphics[width=.96\linewidth]{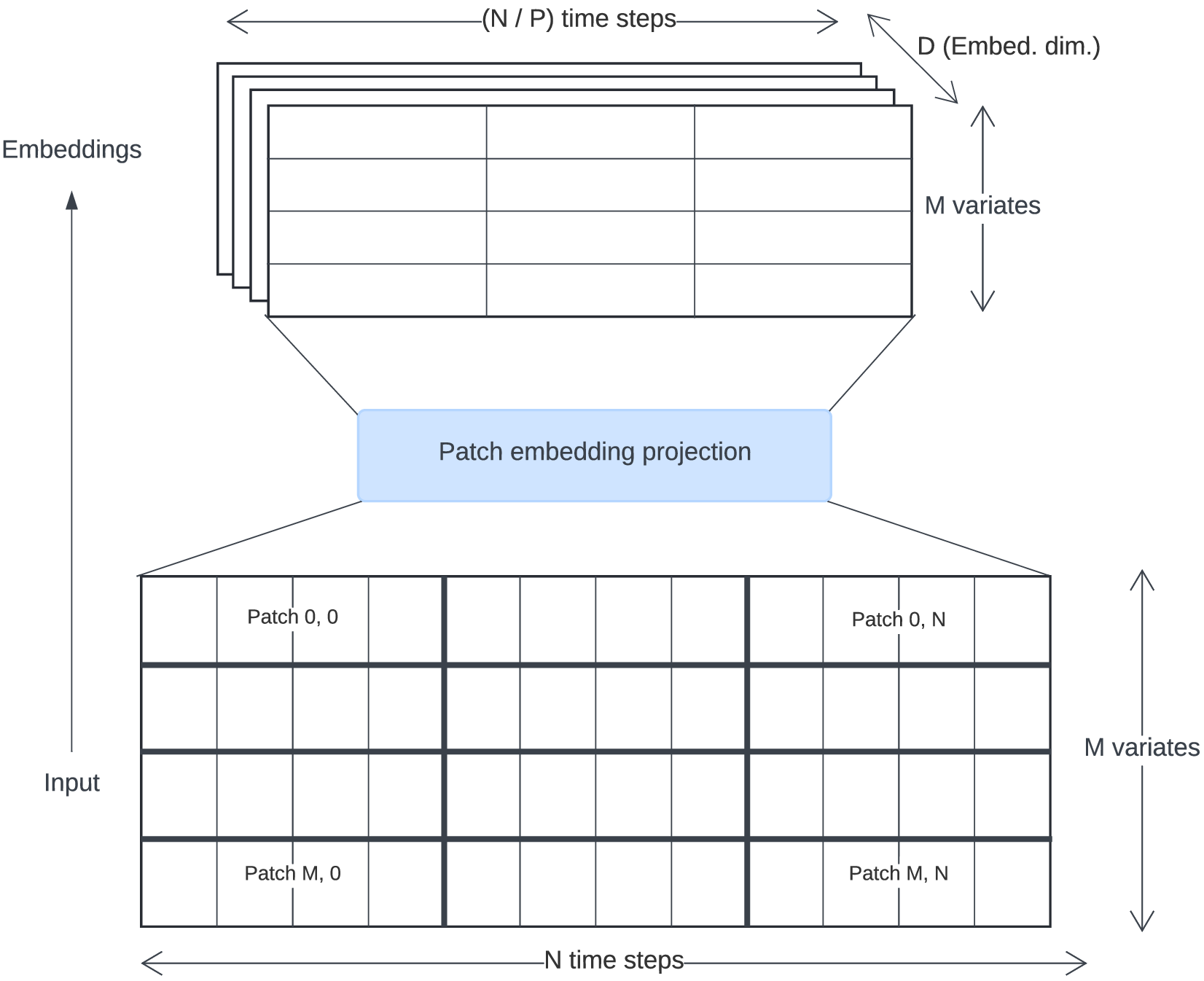}
    \captionof{figure}{The patch embedding takes as input a multivariate time series of \( M \) variates by \( N \) time steps. It divides each variate along the time dimension into patches of size \( P \) and projects these linearly into an embedding space of latent dimension \( D \). This results in an output of size  \( M \times \frac{N}{P} \times D \) which is fed to the transformer decoder.}
    \label{fig:patch_embedding}
\end{minipage}

\subsection{Attention mechanism}
Observability metrics are often high-cardinality, multivariate time series. Therefore, an ideal model will natively handle multivariate forecasting. It should be able to analyze relationships both in the time dimension (what we refer to as “time-wise” interactions) and in the channel dimension (what we refer to as “space-wise” interactions, following the convention in the Datadog platform of describing different groups or tag sets of a metric as the “space” dimension).

In order to model both space and time-wise interactions, we need to adapt the traditional multi-head attention architecture \cite{Vaswani2017} from one to two dimensions. Several approaches have been proposed in the literature to do this, including:

\begin{itemize}
    \item Assuming channel independence, and computing attention only in the time dimension \cite{Nie2023}. This is efficient, but throws away all information about space-wise interactions.
    
    \item Computing attention only in the space dimension, and using a feed-forward network in the time dimension \cite{ilbert2024samformer, Liu2024}.
    
    \item Concatenating variates along the time dimension and computing full cross-attention between every space/time location \cite{Woo2024}. This can capture every possible space and time interaction, but it is computationally costly.
    
    \item Computing “factorized attention,” where each transformer block contains a separate space and time attention computation \cite{zhang2023crossformer, Rao2021, Arnab2021}. This allows both space and time mixing, and is more efficient than full cross-attention. However, it doubles the effective depth of the network.
\end{itemize}

In order to design our attention mechanism, we follow the intuition that for many time series, the time relationships are more important or predictive than the space relationships. As evidence, we observe that even models that completely ignore space-wise relationships (such as PatchTST \cite{Nie2023} and TimesFM \cite{das2024a}) can still achieve competitive performance on multivariate datasets. However, other studies (e.g. Moirai \cite{Woo2024}) have shown through ablations that there is some clear benefit to including space-wise relationships.

We therefore propose a novel variant of factorized attention, which we call “Proportional Factorized Space-Time Attention.” We use a mixture of alternating space-wise and time-wise attention blocks. As a configurable hyperparameter, we can change the ratio of time-wise to space-wise blocks, thus allowing us to devote more or less compute budget to each type of attention. For our base model, we selected a configuration with one space-wise attention block for every two time-wise blocks.

In the time-wise attention blocks, we use causal masking and rotary positional embeddings \cite{su2021roformer} with XPOS \cite{sun2022a} in order to autoregressively model time-dependent features. In the space-wise blocks, by contrast, we use full bidirectional attention in order to preserve permutation invariance of the covariates, with a block-diagonal ID mask to ensure that only related variates attend to each other. This masking allows us to pack multiple independent multivariate time series into the same batch, in order to improve training efficiency and reduce the amount of padding.

\begin{figure*}[!ht]
  \centering
    \includegraphics[width=.85\linewidth]{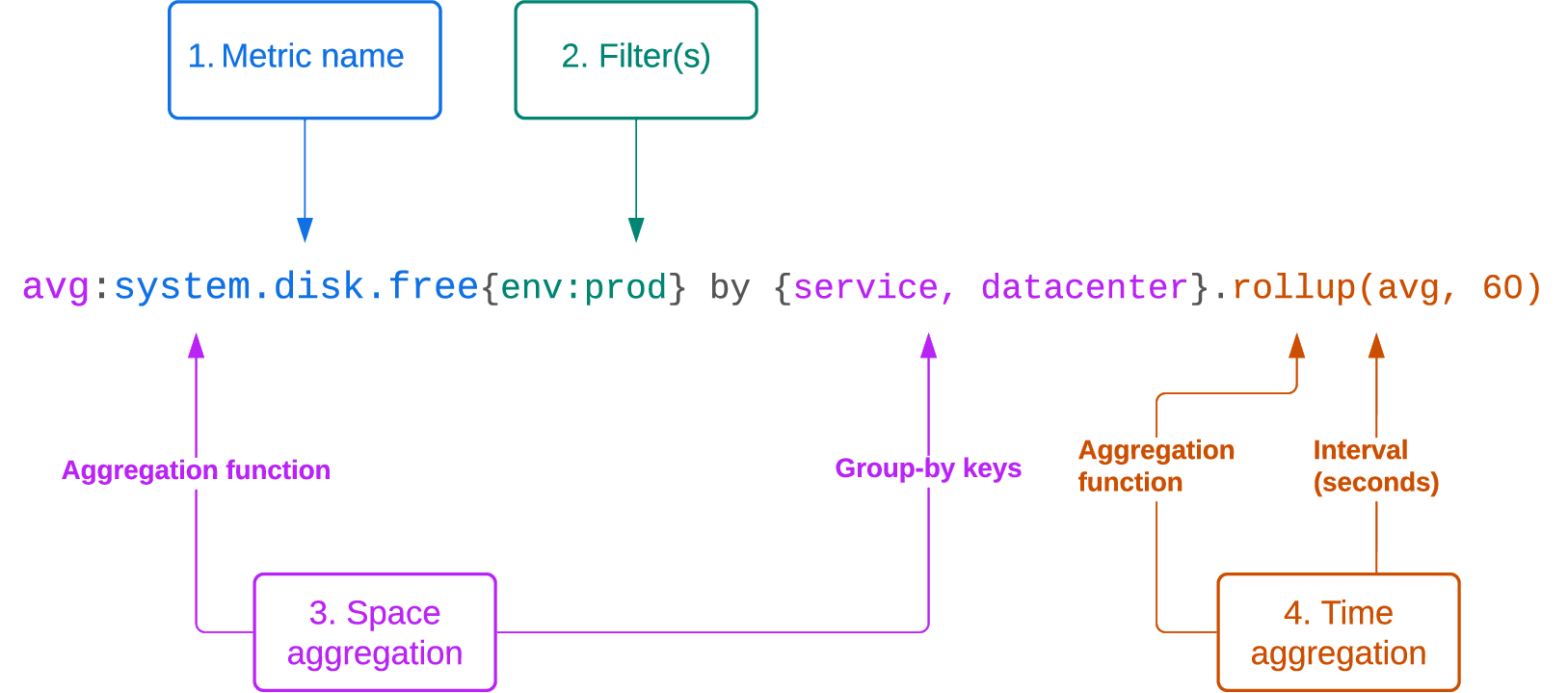}
   \caption{
      Example metric query in the Datadog platform. The metric name (1) determines which metric is being queried. The filter clause (2) limits which contexts are queried, in this case restricting the query to the prod environment. The space aggregation (3) indicates that the average metric value should be returned for each unique combination of the group-by keys. The time aggregation (4) indicates that metric values should be aggregated to the average for each 60-second interval. The query results will be a multivariate time series with 1-minute time steps, and with separate individual variates for each unique {service, datacenter} tuple.
    }
    \label{fig:metric_query}
\end{figure*}

\subsection{Probabilistic prediction head}
In order to be useful for forecasting applications, a model should produce probabilistic predictions. A common practice in time series models is to use an output layer where the model regresses the parameters of a probability distribution. This allows for prediction intervals to be computed using Monte Carlo sampling \cite{Salinas2020}.

Common choices for an output layer are Normal \cite{Salinas2020} and Student-T \cite{das2023longterm, rasul2023lagllama}, which can improve robustness to outliers. Moirai \cite{Woo2024} allows for more flexible residual distributions by proposing a novel mixture model incorporating a weighted combination of Gaussian, Student-T, Log-Normal, and Negative-Binomial outputs.  

However, real-world time series can often have complex distributions that are challenging to fit, with outliers, heavy tails, extreme skew, and multimodality. In order to accommodate these scenarios, we introduce an even more flexible output likelihood. To do this we employ a method based on Gaussian mixture models (GMMs), which can approximate any density function (\cite{Goodfellow-et-al-2016}). To avoid training instability in the presence of outliers, we use  a Student-T mixture model (SMM), a robust generalization of GMMs \cite{Peel2000} that has previously shown promise for modeling heavy-tailed financial time series \cite{Meitz2018AMA, Wong2009}. The model predicts \( k \)  Student-T distributions (where \( k \)  is a hyperparameter) for each time step, as well as a learned weighting.

When we perform inference, we draw samples from the mixture distribution at each timestamp, then feed each sample back into the decoder for the next prediction. This allows us to produce prediction intervals at any quantile, limited only by the number of samples; for more precise tails, we can choose to spend more computation on sampling (\fref{fig:forecast}).

\subsection{Input/output scaling}
As in other time series models, we perform instance normalization on input data before passing it through the patch embedding, in order to make the model generalize better to inputs of different scales \cite{kim2022reversible}. We scale the inputs to have zero mean and unit standard deviation. The output predictions are then rescaled back to the original units.

\subsection{Training objective}
As a decoder-only model, Toto is pre-trained on the next-patch prediction task. We minimize the negative log-likelihood of the next predicted patch with respect to the distribution output of the model. We train the model using the AdamW optimizer \cite{loshchilov2018decoupled}.

\subsection{Hyperparameters}
The hyperparameters used for Toto are detailed in \tref{tab:hyperparameters}, with 103 million total parameters.

\section{Training data} \label{datasets}
We pretrained Toto with a dataset of approximately one trillion time series points. Of these, roughly three-quarters are anonymous observability metrics from the Datadog platform. The remaining points come from the LOTSA dataset \cite{Woo2024}, a compilation of publicly-available time series datasets across many different domains.

\subsection{Datadog dataset}

The Datadog platform ingests more than a hundred  trillion  events per day. However, much of this data is sparse, noisy, or too granular or high in cardinality to be useful in its raw form. To curate a high-quality dataset for efficient model training, we sample queries based on quality and relevance signals from dashboards, monitor alerts, and notebooks. This provides a strong signal that the data resulting from these queries is of critical importance and sufficient quality for observability of real-world applications.

\begin{table*}[!htbp]
  \centering
  \resizebox{\textwidth}{!}{  
    \begin{tabular}{@{}llccccccccccccc@{}}
      \toprule
      & & \multicolumn{5}{c}{\textit{Zero Shot}} & \multicolumn{8}{c}{\textit{Full Shot}} \\
      \cmidrule(lr){3-7} \cmidrule(lr){8-15}
\textbf{Dataset} & \textbf{Metric}  & \textbf{Toto} & \textbf{Moirai\textsubscript{Small}} & \textbf{Moirai\textsubscript{Base}} & \textbf{Moirai\textsubscript{Large}} & \textbf{TimesFM\textsuperscript{*}} & \textbf{iTransformer} & \textbf{TimesNet} & \textbf{PatchTST} & \textbf{Crossformer} & \textbf{TiDE} & \textbf{DLinear} & \textbf{SCINet} & \textbf{FEDformer} \\
      \midrule
      \textbf{ETTh1} & \textbf{MAE} & \textbf{0.389} & \underline{0.424} & 0.438 & 0.469 &  0.426 & 0.448 & 0.450 & 0.455 & 0.522 & 0.507 & 0.452 & 0.647 & 0.460 \\
                     & \textbf{MSE} & \textbf{0.363} & \underline{0.400} & 0.434 & 0.510 &  - & 0.454 & 0.458 & 0.469 & 0.529 & 0.541 & 0.456 & 0.747 & 0.440 \\
        \midrule
      \textbf{ETTh2} & \textbf{MAE} & \textbf{0.261} & 0.379 & 0.382 & \underline{0.376} &  0.410 & 0.407 & 0.497 & 0.407 & 0.684 & 0.550 & 0.515 & 0.723 & 0.449 \\
                     & \textbf{MSE} & \textbf{0.170} & \underline{0.341} & 0.345 & 0.354 &  - & 0.383 & 0.414 & 0.387 & 0.942 & 0.611 & 0.559 & 0.954 & 0.437 \\
        \midrule
      \textbf{ETTm1} & \textbf{MAE} & \textbf{0.375} & 0.409 & \underline{0.388} & 0.389 &  \underline{0.388} & 0.410 & 0.406 & 0.400 & 0.495 & 0.419 & 0.407 & 0.481 & 0.452 \\
                     & \textbf{MSE} & \textbf{0.372} & 0.448 & \underline{0.381} & 0.390 &  - & 0.407 & 0.400 & 0.387 & 0.513 & 0.419 & 0.403 & 0.486 & 0.448 \\
        \midrule
      \textbf{ETTm2} & \textbf{MAE} & \textbf{0.319} & 0.341 & 0.321 & \underline{0.320} &  0.334 & 0.332 & 0.333 & 0.326 & 0.611 & 0.404 & 0.401 & 0.537 & 0.349 \\
                     & \textbf{MSE} & \textbf{0.272} & 0.300 & \textbf{0.272} & \underline{0.276} &   - & 0.288 & 0.291 & 0.281 & 0.757 & 0.358 & 0.350 & 0.571 & 0.305 \\
        \midrule
      \textbf{Electricity} & \textbf{MAE} & \textbf{0.246} & 0.320 & 0.274 & 0.273 &  - & \underline{0.270} & 0.295 & 0.304 & 0.334 & 0.344 & 0.300 & 0.365 & 0.327 \\
                           & \textbf{MSE} & \textbf{0.157} & 0.233 & 0.188 & 0.188 &  - & \underline{0.178} & 0.193 & 0.216 & 0.244 & 0.252 & 0.212 & 0.268 & 0.214 \\
        \midrule
      \textbf{Weather} & \textbf{MAE} & 0.284 & \underline{0.267} & \textbf{0.261} & 0.275 & - & 0.278 & 0.287 & 0.281 & 0.315 & 0.320 & 0.317 & 0.363 & 0.360 \\
                       & \textbf{MSE} & 0.256 & \underline{0.242} & \textbf{0.238} & 0.259 & - & 0.258 & 0.259 & 0.259 & 0.259 & 0.271 & 0.265 & 0.292 & 0.309 \\
      \midrule
      \midrule
      \textbf{Mean} & \textbf{MAE} & \textbf{0.312} & 0.357 & \underline{0.341} & 0.350 & - & 0.357 & 0.378 & 0.362 & 0.493 & 0.424 & 0.399 & 0.519 & 0.400 \\
       & \textbf{MSE} & \textbf{0.265} & 0.328 & \underline{0.315} & 0.330 & - & 0.328 & 0.336 & 0.333 & 0.541 & 0.409 & 0.374 & 0.533 & 0.359 \\
      \bottomrule
    \end{tabular}
  }
  \caption{Comparison of different models with Toto on the LSF benchmark datasets. Results are averaged across prediction lengths of 96, 192, 336, and 720 steps. For Toto, we use a stride of 512 steps and a historical context window of 512 steps. For other models, we use the results reported in \cite{Woo2024} and \cite{das2024a}. Metrics for each  prediction length are available in \tref{tab:lsf_full_prediction_lengths}. \textbf{\textsuperscript{*}}TimesFM only reports values for MAE on ETTh1, ETTh2, ETTm1, and ETTm2. Key: \textbf{Best results}, \underline{Second-best results.}}

  \label{tab:lsf_results} 
\end{table*}

Datadog metrics are accessed using a specialized query language supporting filters, group-bys, time aggregation, and various transformations and postprocessing functions \cite{Datadog_Querying}. We consider groups returned from the same query to be related variates in a multivariate time series (\fref{fig:metric_query}). After we retrieve the query results, we discard the query strings and group identifiers, keeping only the raw numeric data.

Handling this vast amount of data requires several preprocessing steps to ensure consistency and quality. Initially, we apply padding and masking techniques to align the series lengths, making them divisible by the patch stride. This involves adding necessary left-padding to both the time series data and the ID mask, ensuring compatibility with the model\textquotesingle s requirements.

Various data augmentations are employed to enhance the dataset\textquotesingle s robustness. We introduce random time offsets to prevent memorization caused by having series always align the same way with the patch grid. After concatenating the Datadog and LOTSA datasets for training, we also implement a variate shuffling strategy to maintain diversity and representation. Specifically, 10\% of the time, we combine variates that are not necessarily related, thus creating new, diverse combinations of data points. To sample the indices, we employ a normal distribution with a standard deviation of 1000, favoring data points that were closer together in the original datasets. This Gaussian sampling ensures that, while there is a preference for adjacent data points, significant randomness is introduced to enhance the diversity of the training data. This approach improves the model\textquotesingle s ability to generalize across different types of data effectively.

By implementing these rigorous preprocessing steps and sophisticated data handling mechanisms, we ensure that the training data for Toto is of the highest quality, ultimately contributing to the model\textquotesingle s superior performance and robustness.

\subsection{Synthetic data}

We use a  synthetic data generation process similar to TimesFM  \cite{das2024a} to supplement our training datasets, improving the diversity of the data and helping to teach the model basic structure. We simulate time series data through the composition of components such as piecewise linear trends, ARMA processes, sinusoidal seasonal patterns, and various residual distributions. We randomly combine five of these processes per variate, introducing patterns not always present in our real-world datasets. The generation process involves creating base series with random transformations, clipping extreme values, and rescaling to a specified range. By making synthetic data approximately 5\% of our training dataset, we ensure a wide range of time-series behaviors are captured. This diversity exposes our models to various scenarios during training, improving their ability to generalize and effectively handle real-world data.


\section{Results} \label{results}
We report experimental results for a pre-trained Toto model in \sref{lsf_benchmark} and \sref{DD_Benchmark}.

To evaluate predictions, we sequentially divide a time series into context and forecast segments. We input the context segment into Toto and autoregressively generate output patches by sampling from the Student-T mixture model distribution. We forecast a number of steps equal to the nearest multiple of the patch size, then truncate the predictions to the desired length. In order to keep inference time consistent, we vary the number of samples generated based on the cardinality and length of the dataset, with a minimum of 100 samples. We take the median sample at each time step as the final point prediction. This prediction is then compared against the ground-truth forecast segment for evaluation.

\subsection{LSF benchmarks} \label{lsf_benchmark}

To assess general-purpose time series forecasting performance, we use the Long Sequence Forecasting (LSF) benchmark datasets (ETTh1, ETTh2, ETTm1, ETTm2, Electricity, and Weather) \cite{Wu2021}. We evaluate with forecast lengths of 96, 192, 336, and 720 time steps, in sliding windows with stride 512, and average the results. For Toto, we used a historical context window of 512 steps and took the median of 200 samples. Following standard practice, we report normalized Mean Absolute Error (MAE) and Mean Squared Error (MSE), fitted on a training split, in order to be able to compare performance across different datasets. We compared Toto\textquotesingle s performance with the reported results of other recent zero-shot foundation models \cite{Woo2024,das2024a}, as well as full-shot time series forecasting models \cite{Liu2024, wu2023timesnet, Nie2023, zhang2023crossformer, das2023longterm, Zeng_Chen_Zhang_Xu_2023, liu2022scinet, zhou2022fedformer}. We display these results in \tref{tab:lsf_results}.

Toto demonstrates exceptional performance across a variety of benchmark datasets, excelling in zero-shot scenarios. In the LSF datasets, Toto consistently outperforms other models in terms of MAE and MSE. For example, on the ETTh1 dataset, Toto achieves an MAE of 0.389 and an MSE of 0.363, outperforming all zero-shot models, including the previously reported Moirai series and TimesFM. Macro-averaging across the six LSF datasets, Toto achieves an MAE of 0.312 and MSE of 0.265, again exceeding Moirai\textquotesingle s reported zero-shot performance as well as the reported performance of the full-shot models.

Several architectural choices and data features likely contribute to Toto\textquotesingle s superior performance. The novel Proportional Factorized Space-Time Attention mechanism allows Toto to efficiently capture both temporal and spatial dependencies within multivariate time series data. Additionally, the extensive training on a diverse dataset of one trillion time series points, including a mix of real-world observability metrics and multi-domain time series data, enhances Toto\textquotesingle s ability to handle varied characteristics of different benchmark datasets.

While Toto generally excels, there are areas where its performance is closely matched by other models. In full-shot scenarios, models like PatchTST, Crossformer, and FEDformer show competitive results. For example, on the Electricity dataset, while Toto achieves a leading zero-shot MAE of 0.246 and MSE of 0.157, iTransformer and TimesNet also show strong performance, indicating that these models can catch up when additional training data is available.

Overall, Toto\textquotesingle s architectural innovations and extensive training data enable it to achieve state-of-the-art performance across diverse benchmarks, excelling in zero-shot scenarios while remaining highly competitive in full-shot contexts.

\begin{table*}[!ht]
    \centering
    \resizebox{\textwidth}{!}{%
    \begin{tabular}{lcccccccccc}
        \toprule
        \textbf{Metric} & \textbf{Toto} & \textbf{Chronos-T5\textsubscript{Tiny}} & \textbf{Chronos-T5\textsubscript{Mini}} & \textbf{Chronos-T5\textsubscript{Small}} & \textbf{Chronos-T5\textsubscript{Base}} & \textbf{Chronos-T5\textsubscript{Large}} & \textbf{Moirai\textsubscript{Small}} & \textbf{Moirai\textsubscript{Base}} & \textbf{Moirai\textsubscript{Large}} & \textbf{TimesFM} \\
        \midrule
        \textbf{sMAPE} & \textbf{0.672} & 0.809 & 0.788 & 0.800 & 0.796 & 0.805 & 0.808 & 0.742 & \underline{0.736} & 1.246 \\
        \midrule
        \textbf{sMdAPE} & \textbf{0.318} & 0.406 & 0.391 & 0.401 & 0.393 & 0.396 & 0.418 & 0.370 & \underline{0.365} &  0.639 \\
        \bottomrule
    \end{tabular}
    }
    \caption{Performance of Toto and other zero-shot models on the Datadog benchmark dataset. Key: \textbf{Best results}, \underline{Second-best results.}}
    \label{tab:dd_benchmark_results}
\end{table*}

\subsection{Datadog benchmark} \label{DD_Benchmark}

We created a benchmark using anonymous Datadog data to assess performance across various observability metrics. To ensure a representative and realistic sample, we sampled data based on quality and relevance signals from dashboards, monitor alerts, and notebooks. This benchmark comprises 983,994 data points from 82 distinct multivariate time series, encompassing 1,122 variates.

We analyzed summary statistics of the series in our benchmark to identify characteristics that make observability time series challenging to forecast. The categories and their definitions are as follows:

\begin{itemize}
    \item \textbf{Sparse:} Series with a low density of observations, indicating infrequent recording of data or rare events.
    \item \textbf{Extreme right skew:} Series with a distribution heavily skewed to the right, characterized by a few very high values and many lower values.
    \item \textbf{Seasonal:} Series exhibiting regular and recurring patterns, often linked to daily, weekly, or yearly cycles.
    \item \textbf{Flat:} Series with minimal variability, showing little to no change over time.
\end{itemize}

The relative proportion of these cases are displayed in \tref{tab:dd_benchmark_breakdown}.

To assess the prediction of other zero-shot models on the DD Benchmark, we follow sampling procedures delineated in their respective manuscripts. In short, for Chronos models, we generate 20 samples and take the median prediction. For Moirai models, we take the median of 100 samples and set the patch size to “auto”. TimesFM only produces point predictions of the mean, so we use those directly. Since TimesFM and Chronos only support univariate forecasting, we process each variate independently. Moirai, on the other hand, like Toto, makes joint predictions for each group of related variates. For Toto, we utilize the same evaluation procedure we used on the LSF benchmarks.

The evaluation results (\tref{tab:dd_benchmark_results}) demonstrate that Toto outperforms the other models. We evaluate using a prediction length of 365, the maximum forecast window available for previous time series models within the Datadog platform. We use a historical context window of 512 steps. Because observability data can have extreme variation in both magnitude and dispersion, we select symmetric mean absolute percentage error (sMAPE) as a scale-invariant performance metric \cite{armstrong1985long}. We also report symmetric median absolute percentage error (sMdAPE), a robust version of sMAPE \cite{Hyndman2006} that minimizes the influence of the extreme outliers present in observability data. With the lowest sMAPE of 0.672 and sMdAPE of 0.318, Toto proves to be the most accurate for forecasting observability time series data.

These results suggest that current open datasets may not provide sufficient information to extrapolate to the specific nuances of observability data, highlighting the importance of training on more relevant data as demonstrated by Toto\textquotesingle s superior performance.

\begin{minipage}{\linewidth}
    \centering
\makebox[\columnwidth][c]{
\begin{tabular}{lr}
\\
\toprule
Case & \% \\
\midrule
Sparse & 12.20 \\
Extreme Right Skew & 17.07 \\
Seasonal & 80.49 \\
Flat & 1.22 \\
\bottomrule
\end{tabular}
}
\captionof{table}{Breakdown of Datadog dataset based on case, computed based on the average characteristics of variates in each multivariate series. Note that these do not add to 100\% because time series may fall into multiple categories.}
\label{tab:dd_benchmark_breakdown}
\end{minipage}

\section{Conclusions} \label{conclusions}
Toto, through a novel architecture and pre-training corpus, demonstrates state-of-the-art performance both on public benchmarks and on the Datadog observability benchmark. We look forward to sharing many more technical details, experiments, and benchmark results  in a forthcoming paper.

\section{Impact statement} \label{impact_statement}
In developing Toto, Datadog follows a structured approach to ensure responsible development, focusing on identifying, assessing, and mitigating potential risks associated with the use of our model. Given that Toto is not intended for mass distribution and specifically generates time series forecasts for observability data, the potential harms are considerably lower compared to more general-purpose models. At Datadog, our primary focus is on ensuring the accuracy, reliability, and security of the forecasts generated by Toto, which are crucial for maintaining and optimizing infrastructure and application performance.

We carefully analyze the potential for Toto to produce incorrect or misleading forecasts that could impact decision-making processes in critical systems. Additionally, we consider the implications of Toto\textquotesingle s performance across diverse datasets, ensuring it can generalize well without introducing significant errors.

\section{Future directions} \label{future_directions}
Many exciting areas of exploration remain for further study. If you are interested in working with us, please reach out to the authors by email.

Some future research questions that particularly intrigue us include:

\begin{itemize}
    \item \textbf{Multi-modal inputs:} Incorporate additional input modalities such as query metadata and captions to enhance forecast performance.

    \item \textbf{Autonomous troubleshooting agents:} Augment Datadog\textquotesingle s AI agents \cite{Datadog_Bits_AI} for troubleshooting and incident response by integrating modality-specific foundation models like Toto to improve their reasoning and planning capabilities with telemetry data.

    \item \textbf{Conversational interfaces:} Align time series forecasting models with LLMs to develop conversational agents capable of interpreting and reasoning about time series data.

    \item  \textbf{Model enhancements and scaling:} Explore ways to improve and scale model performance through optimizations such as new types of input embeddings, attention mechanisms, and examining alternative variate groupings to capture richer interactions.

\end{itemize}

\section{Contributions} \label{contributions}
\textit{Contributors are listed in alphabetical order.}

Othmane Abou-Amal, Joseph Banks, Mayeul Blanzat, Ben Cohen, Youssef Doubli, Ben Hinthorne, Emaad Khwaja, Jared Ledvina, Charles Masson, Sajid Mehmood, Elise Ramé, Maxime Visonneau, Kan Wang.

\section{Acknowledgements} \label{acknowledgements}
Our work is made possible by the efforts of numerous teams at Datadog. Special thanks and acknowledgement to:  

Johan Andersen, Roashan Ayene, Romoli Bakshi, Kevin Beach, Bill Birkholz, Rob Boll, Maxim Brown, Benedetto Buratti, Marion Chan-Renous, Jessica Cordonnier, Ben Donohue, Zakaria Fikrat, Quentin François, Erica Hale, Michael Hoang, Joe Jones, Max Livingston, Jesse Mack, Amine Naouas, Sean O\textquotesingle Connor, Brendan Rhoads, Phil Sarin, Vyom Shah, Aaron Taa, Bharath Vontimitta, Dominique West, Steven Zhou.

{\scriptsize
\bibliographystyle{unsrtnat}
\bibliography{references}

\begin{thebibliography}{50}
\providecommand{\natexlab}[1]{#1}
\providecommand{\url}[1]{\texttt{#1}}
\expandafter\ifx\csname urlstyle\endcsname\relax
  \providecommand{\doi}[1]{doi: #1}\else
  \providecommand{\doi}{doi: \begingroup \urlstyle{rm}\Url}\fi

\bibitem[Datadog(2024{\natexlab{a}})]{Datadog_Observability}
Datadog.
\newblock Observability platform, 2024{\natexlab{a}}.
\newblock URL \url{https://www.datadoghq.com/monitoring/observability-platform/}.

\bibitem[Datadog(2024{\natexlab{b}})]{Datadog_Infrastructure}
Datadog.
\newblock Modern infrastructure monitoring, 2024{\natexlab{b}}.
\newblock URL \url{https://www.datadoghq.com/product/infrastructure-monitoring/}.

\bibitem[Hyndman and Athanasopoulos(2021)]{Hyndman2021}
Rob~J Hyndman and George Athanasopoulos.
\newblock \emph{Forecasting: Principles and Practice}.
\newblock OTexts, 3rd edition, 2021.
\newblock URL \url{https://otexts.com/fpp3/}.

\bibitem[Fildes et~al.(1998)Fildes, Hibon, Makridakis, and Meade]{Fildes1998}
Robert Fildes, Michèle Hibon, Spyros Makridakis, and Nigel Meade.
\newblock Generalising about univariate forecasting methods: further empirical evidence.
\newblock \emph{International Journal of Forecasting}, 14:\penalty0 339--358, 9 1998.
\newblock ISSN 01692070.
\newblock \doi{10.1016/S0169-2070(98)00009-0}.

\bibitem[Stevenson(2007)]{Stevenson2007}
Simon Stevenson.
\newblock A comparison of the forecasting ability of arima models.
\newblock \emph{Journal of Property Investment \& Finance}, 25:\penalty0 223--240, 5 2007.
\newblock ISSN 1463-578X.
\newblock \doi{10.1108/14635780710746902}.

\bibitem[Christodoulos et~al.(2010)Christodoulos, Michalakelis, and Varoutas]{Christodoulos2010}
Charisios Christodoulos, Christos Michalakelis, and Dimitris Varoutas.
\newblock Forecasting with limited data: Combining arima and diffusion models.
\newblock \emph{Technological Forecasting and Social Change}, 77:\penalty0 558--565, 5 2010.
\newblock ISSN 00401625.
\newblock \doi{10.1016/j.techfore.2010.01.009}.

\bibitem[Salinas et~al.(2020)Salinas, Flunkert, Gasthaus, and Januschowski]{Salinas2020}
David Salinas, Valentin Flunkert, Jan Gasthaus, and Tim Januschowski.
\newblock Deepar: Probabilistic forecasting with autoregressive recurrent networks.
\newblock \emph{International Journal of Forecasting}, 36:\penalty0 1181--1191, 2020.
\newblock ISSN 0169-2070.
\newblock \doi{https://doi.org/10.1016/j.ijforecast.2019.07.001}.
\newblock URL \url{https://www.sciencedirect.com/science/article/pii/S0169207019301888}.

\bibitem[Brophy et~al.(2023)Brophy, Wang, She, and Ward]{Brophy2023}
Eoin Brophy, Zhengwei Wang, Qi~She, and Tomás Ward.
\newblock Generative adversarial networks in time series: A systematic literature review.
\newblock \emph{ACM Computing Surveys}, 55:\penalty0 1--31, 10 2023.
\newblock ISSN 0360-0300.
\newblock \doi{10.1145/3559540}.

\bibitem[Jia et~al.(2018)Jia, Lin, Qi, and Aiken]{Jia2018}
Zhihao Jia, Sina Lin, Charles~R Qi, and Alex Aiken.
\newblock Exploring the hidden dimension in accelerating convolutional neural networks, 2018.
\newblock URL \url{https://openreview.net/forum?id=SJCPLLpaW}.

\bibitem[Xu et~al.(2021)Xu, Zhang, and Tang]{Xu2021}
Weizheng Xu, Youtao Zhang, and Xulong Tang.
\newblock Parallelizing dnn training on gpus: Challenges and opportunities.
\newblock pages 174--178. ACM, 4 2021.
\newblock ISBN 9781450383134.
\newblock \doi{10.1145/3442442.3452055}.

\bibitem[Vaswani et~al.(2017)Vaswani, Shazeer, Parmar, Uszkoreit, Jones, Gomez, Łukasz Kaiser, and Polosukhin]{Vaswani2017}
Ashish Vaswani, Noam Shazeer, Niki Parmar, Jakob Uszkoreit, Llion Jones, Aidan~N Gomez, Łukasz Kaiser, and Illia Polosukhin.
\newblock Attention is all you need.
\newblock volume~30. Curran Associates, Inc., 2017.
\newblock URL \url{https://papers.nips.cc/paper_files/paper/2017/hash/3f5ee243547dee91fbd053c1c4a845aa-Abstract.html}.

\bibitem[Wu et~al.(2021)Wu, Xu, Wang, and Long]{Wu2021}
Haixu Wu, Jiehui Xu, Jianmin Wang, and Mingsheng Long.
\newblock Autoformer: Decomposition transformers with auto-correlation for long-term series forecasting.
\newblock 2021.
\newblock URL \url{https://openreview.net/forum?id=J4gRj6d5Qm}.

\bibitem[Zhou et~al.(2020)Zhou, Zhang, Peng, Zhang, Li, Xiong, and Zhang]{Zhou2020}
Haoyi Zhou, Shanghang Zhang, Jieqi Peng, Shuai Zhang, Jianxin Li, Hui Xiong, and Wan Zhang.
\newblock Informer: Beyond efficient transformer for long sequence time-series forecasting.
\newblock 2020.
\newblock URL \url{https://api.semanticscholar.org/CorpusID:229156802}.

\bibitem[Nie et~al.(2023)Nie, Nguyen, Sinthong, and Kalagnanam]{Nie2023}
Yuqi Nie, Nam~H Nguyen, Phanwadee Sinthong, and Jayant Kalagnanam.
\newblock A time series is worth 64 words: Long-term forecasting with transformers.
\newblock 2023.
\newblock URL \url{https://openreview.net/forum?id=Jbdc0vTOcol}.

\bibitem[Woo et~al.(2024)Woo, Liu, Kumar, Xiong, Savarese, and Sahoo]{Woo2024}
Gerald Woo, Chenghao Liu, Akshat Kumar, Caiming Xiong, Silvio Savarese, and Doyen Sahoo.
\newblock Unified training of universal time series forecasting transformers.
\newblock 2024.
\newblock URL \url{https://openreview.net/forum?id=Yd8eHMY1wz}.

\bibitem[Zhang and Yan(2023)]{zhang2023crossformer}
Yunhao Zhang and Junchi Yan.
\newblock Crossformer: Transformer utilizing cross-dimension dependency for multivariate time series forecasting.
\newblock In \emph{The Eleventh International Conference on Learning Representations}, 2023.
\newblock URL \url{https://openreview.net/forum?id=vSVLM2j9eie}.

\bibitem[Liu et~al.(2024)Liu, Hu, Zhang, Wu, Wang, Ma, and Long]{Liu2024}
Yong Liu, Tengge Hu, Haoran Zhang, Haixu Wu, Shiyu Wang, Lintao Ma, and Mingsheng Long.
\newblock itransformer: Inverted transformers are effective for time series forecasting.
\newblock 2024.
\newblock URL \url{https://openreview.net/forum?id=JePfAI8fah}.

\bibitem[Ilbert et~al.(2024)Ilbert, Odonnat, Feofanov, Virmaux, Paolo, Palpanas, and Redko]{ilbert2024samformer}
Romain Ilbert, Ambroise Odonnat, Vasilii Feofanov, Aladin Virmaux, Giuseppe Paolo, Themis Palpanas, and Ievgen Redko.
\newblock {SAM}former: Unlocking the potential of transformers in time series forecasting with sharpness-aware minimization and channel-wise attention.
\newblock In \emph{Forty-first International Conference on Machine Learning}, 2024.
\newblock URL \url{https://openreview.net/forum?id=8kLzL5QBh2}.

\bibitem[Das et~al.(2024)Das, Kong, Sen, and Zhou]{das2024a}
Abhimanyu Das, Weihao Kong, Rajat Sen, and Yichen Zhou.
\newblock A decoder-only foundation model for time-series forecasting.
\newblock In \emph{Forty-first International Conference on Machine Learning}, 2024.
\newblock URL \url{https://openreview.net/forum?id=jn2iTJas6h}.

\bibitem[Lin et~al.(2021)Lin, Wang, Liu, and Qiu]{Lin_2021_Survey}
Tianyang Lin, Yuxin Wang, Xiangyang Liu, and Xipeng Qiu.
\newblock A survey of transformers.
\newblock \emph{CoRR}, abs/2106.04554, 2021.
\newblock URL \url{https://arxiv.org/abs/2106.04554}.

\bibitem[Ansari et~al.(2024)Ansari, Stella, Turkmen, Zhang, Mercado, Shen, Shchur, Rangapuram, Arango, Kapoor, Zschiegner, Maddix, Wang, Mahoney, Torkkola, Wilson, Bohlke-Schneider, and Wang]{ansari2024chronoslearninglanguagetime}
Abdul~Fatir Ansari, Lorenzo Stella, Caner Turkmen, Xiyuan Zhang, Pedro Mercado, Huibin Shen, Oleksandr Shchur, Syama~Sundar Rangapuram, Sebastian~Pineda Arango, Shubham Kapoor, Jasper Zschiegner, Danielle~C. Maddix, Hao Wang, Michael~W. Mahoney, Kari Torkkola, Andrew~Gordon Wilson, Michael Bohlke-Schneider, and Yuyang Wang.
\newblock Chronos: Learning the language of time series, 2024.
\newblock URL \url{https://arxiv.org/abs/2403.07815}.

\bibitem[Garza and Mergenthaler-Canseco(2023)]{garza2023timegpt1}
Azul Garza and Max Mergenthaler-Canseco.
\newblock Timegpt-1, 2023.

\bibitem[Rasul et~al.(2023)Rasul, Ashok, Williams, Khorasani, Adamopoulos, Bhagwatkar, Bilo{\v{s}}, Ghonia, Hassen, Schneider, Garg, Drouin, Chapados, Nevmyvaka, and Rish]{rasul2023lagllama}
Kashif Rasul, Arjun Ashok, Andrew~Robert Williams, Arian Khorasani, George Adamopoulos, Rishika Bhagwatkar, Marin Bilo{\v{s}}, Hena Ghonia, Nadhir Hassen, Anderson Schneider, Sahil Garg, Alexandre Drouin, Nicolas Chapados, Yuriy Nevmyvaka, and Irina Rish.
\newblock Lag-llama: Towards foundation models for time series forecasting.
\newblock In \emph{R0-FoMo:Robustness of Few-shot and Zero-shot Learning in Large Foundation Models}, 2023.
\newblock URL \url{https://openreview.net/forum?id=jYluzCLFDM}.

\bibitem[Gruver et~al.(2023)Gruver, Finzi, Qiu, and Wilson]{gruver2023large}
Nate Gruver, Marc~Anton Finzi, Shikai Qiu, and Andrew~Gordon Wilson.
\newblock Large language models are zero-shot time series forecasters.
\newblock In \emph{Thirty-seventh Conference on Neural Information Processing Systems}, 2023.
\newblock URL \url{https://openreview.net/forum?id=md68e8iZK1}.

\bibitem[Radford and Narasimhan(2018)]{Radford2018ImprovingLU}
Alec Radford and Karthik Narasimhan.
\newblock Improving language understanding by generative pre-training.
\newblock 2018.
\newblock URL \url{https://api.semanticscholar.org/CorpusID:49313245}.

\bibitem[Radford et~al.(2019)Radford, Wu, Child, Luan, Amodei, and Sutskever]{Radford2019LanguageMA}
Alec Radford, Jeff Wu, Rewon Child, David Luan, Dario Amodei, and Ilya Sutskever.
\newblock Language models are unsupervised multitask learners.
\newblock 2019.
\newblock URL \url{https://api.semanticscholar.org/CorpusID:160025533}.

\bibitem[Xiong et~al.(2020)Xiong, Yang, He, Zheng, Zheng, Zhang, Lan, Wang, and Liu]{xiong2020on}
Ruibin Xiong, Yunchang Yang, Di~He, Kai Zheng, Shuxin Zheng, Huishuai Zhang, Yanyan Lan, Liwei Wang, and Tie-Yan Liu.
\newblock On layer normalization in the transformer architecture, 2020.
\newblock URL \url{https://openreview.net/forum?id=B1x8anVFPr}.

\bibitem[Zhang and Sennrich(2019)]{zhang-sennrich-neurips19}
Biao Zhang and Rico Sennrich.
\newblock {Root Mean Square Layer Normalization}.
\newblock In \emph{Advances in Neural Information Processing Systems 32}, Vancouver, Canada, 2019.
\newblock URL \url{https://openreview.net/references/pdf?id=S1qBAf6rr}.

\bibitem[Shazeer(2020)]{shazeer2020gluvariantsimprovetransformer}
Noam Shazeer.
\newblock Glu variants improve transformer, 2020.
\newblock URL \url{https://arxiv.org/abs/2002.05202}.

\bibitem[Cordonnier et~al.(2020)Cordonnier, Loukas, and Jaggi]{DBLP:conf/iclr/CordonnierLJ20}
Jean{-}Baptiste Cordonnier, Andreas Loukas, and Martin Jaggi.
\newblock On the relationship between self-attention and convolutional layers.
\newblock In \emph{8th International Conference on Learning Representations, {ICLR} 2020, Addis Ababa, Ethiopia, April 26-30, 2020}. OpenReview.net, 2020.
\newblock URL \url{https://openreview.net/forum?id=HJlnC1rKPB}.

\bibitem[Dosovitskiy et~al.(2021)Dosovitskiy, Beyer, Kolesnikov, Weissenborn, Zhai, Unterthiner, Dehghani, Minderer, Heigold, Gelly, Uszkoreit, and Houlsby]{dosovitskiy2021an}
Alexey Dosovitskiy, Lucas Beyer, Alexander Kolesnikov, Dirk Weissenborn, Xiaohua Zhai, Thomas Unterthiner, Mostafa Dehghani, Matthias Minderer, Georg Heigold, Sylvain Gelly, Jakob Uszkoreit, and Neil Houlsby.
\newblock An image is worth 16x16 words: Transformers for image recognition at scale.
\newblock In \emph{International Conference on Learning Representations}, 2021.
\newblock URL \url{https://openreview.net/forum?id=YicbFdNTTy}.

\bibitem[Rao et~al.(2021)Rao, Liu, Verkuil, Meier, Canny, Abbeel, Sercu, and Rives]{Rao2021}
Roshan~M Rao, Jason Liu, Robert Verkuil, Joshua Meier, John Canny, Pieter Abbeel, Tom Sercu, and Alexander Rives.
\newblock Msa transformer.
\newblock In Marina Meila and Tong Zhang, editors, \emph{Proceedings of the 38th International Conference on Machine Learning}, volume 139 of \emph{Proceedings of Machine Learning Research}, pages 8844--8856. PMLR, 18--24 Jul 2021.
\newblock URL \url{https://proceedings.mlr.press/v139/rao21a.html}.

\bibitem[Arnab et~al.(2021)Arnab, Dehghani, Heigold, Sun, Lučić, and Schmid]{Arnab2021}
Anurag Arnab, Mostafa Dehghani, Georg Heigold, Chen Sun, Mario Lučić, and Cordelia Schmid.
\newblock Vivit: A video vision transformer.
\newblock In \emph{2021 IEEE/CVF International Conference on Computer Vision (ICCV)}, pages 6816--6826, 2021.
\newblock \doi{10.1109/ICCV48922.2021.00676}.

\bibitem[Su et~al.(2021)Su, Lu, Pan, Wen, and Liu]{su2021roformer}
Jianlin Su, Yu~Lu, Shengfeng Pan, Bo~Wen, and Yunfeng Liu.
\newblock Roformer: Enhanced transformer with rotary position embedding, 2021.

\bibitem[Sun et~al.(2022)Sun, Dong, Patra, Ma, Huang, Benhaim, Chaudhary, Song, and Wei]{sun2022a}
Yutao Sun, Li~Dong, Barun Patra, Shuming Ma, Shaohan Huang, Alon Benhaim, Vishrav Chaudhary, Xia Song, and Furu Wei.
\newblock A length-extrapolatable transformer.
\newblock In \emph{ACL 2023}, December 2022.
\newblock URL \url{https://www.microsoft.com/en-us/research/publication/a-length-extrapolatable-transformer/}.

\bibitem[Das et~al.(2023)Das, Kong, Leach, Mathur, Sen, and Yu]{das2023longterm}
Abhimanyu Das, Weihao Kong, Andrew Leach, Shaan~K Mathur, Rajat Sen, and Rose Yu.
\newblock Long-term forecasting with ti{DE}: Time-series dense encoder.
\newblock \emph{Transactions on Machine Learning Research}, 2023.
\newblock ISSN 2835-8856.
\newblock URL \url{https://openreview.net/forum?id=pCbC3aQB5W}.

\bibitem[Goodfellow et~al.(2016)Goodfellow, Bengio, and Courville]{Goodfellow-et-al-2016}
Ian Goodfellow, Yoshua Bengio, and Aaron Courville.
\newblock \emph{Deep Learning}.
\newblock MIT Press, 2016.
\newblock \url{http://www.deeplearningbook.org}.

\bibitem[Peel and McLachlan(2000)]{Peel2000}
D.~Peel and G.J. McLachlan.
\newblock Robust mixture modelling using the t distribution.
\newblock \emph{Statistics and Computing}, 10\penalty0 (4):\penalty0 339--348, 2000.

\bibitem[Meitz et~al.(2018)Meitz, Preve, and Saikkonen]{Meitz2018AMA}
Mika Meitz, Daniel P.~A. Preve, and Pentti Saikkonen.
\newblock A mixture autoregressive model based on student’s t–distribution.
\newblock \emph{Communications in Statistics - Theory and Methods}, 52:\penalty0 499 -- 515, 2018.
\newblock URL \url{https://api.semanticscholar.org/CorpusID:73615847}.

\bibitem[WONG et~al.(2009)WONG, CHAN, and KAM]{Wong2009}
C.~S. WONG, W.~S. CHAN, and P.~L. KAM.
\newblock A student t -mixture autoregressive model with applications to heavy-tailed financial data.
\newblock \emph{Biometrika}, 96\penalty0 (3):\penalty0 751--760, 2009.
\newblock ISSN 00063444, 14643510.
\newblock URL \url{http://www.jstor.org/stable/27798861}.

\bibitem[Kim et~al.(2022)Kim, Kim, Tae, Park, Choi, and Choo]{kim2022reversible}
Taesung Kim, Jinhee Kim, Yunwon Tae, Cheonbok Park, Jang-Ho Choi, and Jaegul Choo.
\newblock Reversible instance normalization for accurate time-series forecasting against distribution shift.
\newblock In \emph{International Conference on Learning Representations}, 2022.
\newblock URL \url{https://openreview.net/forum?id=cGDAkQo1C0p}.

\bibitem[Loshchilov and Hutter(2019)]{loshchilov2018decoupled}
Ilya Loshchilov and Frank Hutter.
\newblock Decoupled weight decay regularization.
\newblock In \emph{International Conference on Learning Representations}, 2019.
\newblock URL \url{https://openreview.net/forum?id=Bkg6RiCqY7}.

\bibitem[Datadog(2024{\natexlab{c}})]{Datadog_Querying}
Datadog.
\newblock Querying, 2024{\natexlab{c}}.
\newblock URL \url{https://docs.datadoghq.com/dashboards/querying/}.

\bibitem[Wu et~al.(2023)Wu, Hu, Liu, Zhou, Wang, and Long]{wu2023timesnet}
Haixu Wu, Tengge Hu, Yong Liu, Hang Zhou, Jianmin Wang, and Mingsheng Long.
\newblock Timesnet: Temporal 2d-variation modeling for general time series analysis.
\newblock In \emph{International Conference on Learning Representations}, 2023.

\bibitem[Zeng et~al.(2023)Zeng, Chen, Zhang, and Xu]{Zeng_Chen_Zhang_Xu_2023}
Ailing Zeng, Muxi Chen, Lei Zhang, and Qiang Xu.
\newblock Are transformers effective for time series forecasting?
\newblock \emph{Proceedings of the AAAI Conference on Artificial Intelligence}, 37\penalty0 (9):\penalty0 11121--11128, Jun. 2023.
\newblock \doi{10.1609/aaai.v37i9.26317}.
\newblock URL \url{https://ojs.aaai.org/index.php/AAAI/article/view/26317}.

\bibitem[LIU et~al.(2022)LIU, Zeng, Chen, Xu, LAI, Ma, and Xu]{liu2022scinet}
Minhao LIU, Ailing Zeng, Muxi Chen, Zhijian Xu, Qiuxia LAI, Lingna Ma, and Qiang Xu.
\newblock {SCIN}et: Time series modeling and forecasting with sample convolution and interaction.
\newblock In Alice~H. Oh, Alekh Agarwal, Danielle Belgrave, and Kyunghyun Cho, editors, \emph{Advances in Neural Information Processing Systems}, 2022.
\newblock URL \url{https://openreview.net/forum?id=AyajSjTAzmg}.

\bibitem[Zhou et~al.(2022)Zhou, Ma, Wen, Wang, Sun, and Jin]{zhou2022fedformer}
Tian Zhou, Ziqing Ma, Qingsong Wen, Xue Wang, Liang Sun, and Rong Jin.
\newblock {FEDformer}: Frequency enhanced decomposed transformer for long-term series forecasting.
\newblock In \emph{Proc. 39th International Conference on Machine Learning (ICML 2022)}, 2022.

\bibitem[Armstrong(1985)]{armstrong1985long}
J.~Scott Armstrong.
\newblock \emph{Long-range Forecasting: From Crystal Ball to Computer}.
\newblock John Wiley \& Sons, New York, 1985.
\newblock ISBN 9780471822608.

\bibitem[Hyndman and Koehler(2006)]{Hyndman2006}
R.~J Hyndman and A.~B. Koehler.
\newblock Another look at measures of forecast accuracy.
\newblock \emph{International Journal of Forecasting}, 22, 2006.

\bibitem[Datadog(2024{\natexlab{d}})]{Datadog_Bits_AI}
Datadog.
\newblock Bits ai: Reimagining the way you run operations with autonomous investigations, 2024{\natexlab{d}}.
\newblock URL \url{https://www.datadoghq.com/blog/bits-ai-autonomous-investigations}.

\end{thebibliography}
}

\end{multicols}

\newpage
\appendix
\section{Model architecture}
\begin{table*}[!hbp]
    \centering
    \begin{tabular}{lr}
        \toprule
        \textbf{Hyperparameter} & \textbf{Value} \\
        \midrule
        Embedding Dimension & 512 \\
        MLP Dimension & 2048 \\
        \# Layers & 24 \\
        \# Heads & 8 \\
        \# Variates & 32 \\
        (\( \beta_1\), \kern0.5em \( \beta_2\)) & (0.9, 0.95)\\
        Weight Decay & 0.01 \\
        Spacewise Layer Cadence & 3 \\
        Patch Size & 32 \\
        \# Student-T Mixture Model Components & 16 \\
        Initial Learning Rate & 0.001 \\
        Annealing Schedule & Cosine \\
        Batch Size & 192 \\
        Warmup Steps & 5000 \\
        Total Train Steps & 193000 \\
        \bottomrule
    \end{tabular}
    \caption{Hyperparameters for Toto}
    \label{tab:hyperparameters}
\end{table*}

\newpage
\section{Results}
\begin{table}[htbp]
\resizebox{\textwidth}{!}{%
\begin{tabular}{ccccccccccccccccc}
\toprule
\multicolumn{3}{c}{} & \multicolumn{5}{c}{\textit{Zero Shot}} & \multicolumn{8}{c}{\textit{Full Shot}} \\
\cmidrule(lr){4-8} \cmidrule(lr){9-16}
\textbf{Dataset} & \textbf{Prediction Length}  & \textbf{Metric}  & \textbf{Toto} & \textbf{Moirai\textsubscript{Small}} & \textbf{Moirai\textsubscript{Base}} & \textbf{Moirai\textsubscript{Large}} & \textbf{TimesFM} & \textbf{iTransformer} & \textbf{TimesNet} & \textbf{PatchTST} & \textbf{Crossformer} & \textbf{TiDE} & \textbf{DLinear} & \textbf{SCINet} & \textbf{FEDformer} \\
\midrule
    ~ & \textbf{96} & \textbf{MAE} & \textbf{0.366} & 0.402 & 0.402 & \underline{0.398} & \underline{0.398} & 0.405 & 0.402 & 0.419 & 0.448 & 0.464 & 0.400 & 0.599 & 0.419 \\ 
    ~ & ~ & \textbf{MSE} & \textbf{0.307} & \underline{0.375} & 0.384 & 0.380 & - & 0.386 & 0.384 & 0.414 & 0.423 & 0.479 & 0.386 & 0.654 & 0.376 \\ 
    ~ & \textbf{192} & \textbf{MAE} & \textbf{0.368} & \underline{0.419} & 0.429 & 0.434 & 0.424 & 0.436 & 0.429 & 0.445 & 0.474 & 0.492 & 0.432 & 0.631 & 0.448 \\ 
    \textbf{ETTh1} & ~ & \textbf{MSE} & \textbf{0.329} & \underline{0.399} & 0.425 & 0.440 & - & 0.441 & 0.436 & 0.460 & 0.471 & 0.525 & 0.437 & 0.719 & 0.420 \\ 
    ~ & \textbf{336} & \textbf{MAE} & \textbf{0.399} & \underline{0.429} & 0.450 & 0.474 & 0.436 & 0.458 & 0.469 & 0.466 & 0.546 & 0.515 & 0.459 & 0.659 & 0.465 \\ 
    ~ & ~ & \textbf{MSE} & \textbf{0.396} & \underline{0.412} & 0.456 & 0.514 & - & 0.487 & 0.491 & 0.501 & 0.570 & 0.565 & 0.481 & 0.778 & 0.459 \\ 
    ~ & \textbf{720} & \textbf{MAE} & \textbf{0.424} & \underline{0.444} & 0.473 & 0.568 & 0.445 & 0.491 & 0.500 & 0.488 & 0.621 & 0.558 & 0.516 & 0.699 & 0.507 \\ 
    ~ & ~ & \textbf{MSE} & \underline{0.419} & \textbf{0.413} & 0.470 & 0.705 & - & 0.503 & 0.521 & 0.500 & 0.653 & 0.594 & 0.519 & 0.836 & 0.506 \\ 
    \midrule
    ~ & \textbf{96} & \textbf{MAE} & \textbf{0.197} & 0.334 & 0.327 & \underline{0.325} & 0.356 & 0.349 & 0.374 & 0.348 & 0.584 & 0.440 & 0.387 & 0.621 & 0.397 \\ 
    ~ & ~ & \textbf{MSE} & \textbf{0.093} & 0.281 & \underline{0.277} & 0.287 & - & 0.297 & 0.340 & 0.302 & 0.745 & 0.400 & 0.333 & 0.707 & 0.358 \\ 
    ~ & \textbf{192} & \textbf{MAE} & \textbf{0.231} & 0.373 & 0.374 & \underline{0.367} & 0.400 & 0.400 & 0.414 & 0.400 & 0.656 & 0.509 & 0.476 & 0.689 & 0.439 \\ 
    \textbf{ETTh2} & ~ & \textbf{MSE} & \textbf{0.135} & \underline{0.340} & \underline{0.340} & 0.347 & - & 0.380 & 0.402 & 0.388 & 0.877 & 0.528 & 0.477 & 0.860 & 0.429 \\ 
    ~ & \textbf{336} & \textbf{MAE} & \textbf{0.260} & \underline{0.393} & 0.401 & \underline{0.393} & 0.428 & 0.432 & 0.541 & 0.433 & 0.731 & 0.571 & 0.541 & 0.744 & 0.487 \\ 
    ~ & ~ & \textbf{MSE} & \textbf{0.160} & \underline{0.362} & 0.371 & 0.377 & - & 0.428 & 0.452 & 0.426 & 1.043 & 0.643 & 0.594 & 1.000 & 0.496 \\ 
    ~ & \textbf{720} & \textbf{MAE} & \textbf{0.355} & \underline{0.416} & 0.426 & 0.421 & 0.457 & 0.445 & 0.657 & 0.446 & 0.763 & 0.679 & 0.657 & 0.838 & 0.474 \\ 
    ~ & ~ & \textbf{MSE} & \textbf{0.294} & \underline{0.380} & 0.394 & 0.404 & - & 0.427 & 0.462 & 0.431 & 1.104 & 0.874 & 0.831 & 1.249 & 0.463 \\ 
    \midrule
    ~ & \textbf{96} & \textbf{MAE} & \textbf{0.328} & 0.383 & 0.360 & 0.363 & \underline{0.345} & 0.368 & 0.375 & 0.367 & 0.426 & 0.387 & 0.372 & 0.438 & 0.419 \\ 
    ~ & ~ & \textbf{MSE} & \textbf{0.306} & 0.404 & 0.335 & 0.353 & - & 0.334 & 0.338 & \underline{0.329} & 0.404 & 0.364 & 0.345 & 0.418 & 0.379 \\ 
    ~ & \textbf{192} & \textbf{MAE} & \textbf{0.353} & 0.402 & 0.379 & 0.380 & \underline{0.374} & 0.391 & 0.387 & 0.385 & 0.451 & 0.404 & 0.389 & 0.450 & 0.441 \\ 
    \textbf{ETTm1} & ~ & \textbf{MSE} & \textbf{0.328} & 0.435 & \underline{0.366} & 0.376 & - & 0.377 & 0.374 & 0.367 & 0.450 & 0.398 & 0.380 & 0.439 & 0.426 \\ 
    ~ & \textbf{336} & \textbf{MAE} & \textbf{0.389} & 0.416 & \underline{0.394} & 0.395 & 0.397 & 0.420 & 0.411 & 0.410 & 0.515 & 0.425 & 0.413 & 0.485 & 0.459 \\ 
    ~ & ~ & \textbf{MSE} & \textbf{0.390} & 0.462 & \underline{0.391} & 0.399 & - & 0.426 & 0.410 & 0.399 & 0.532 & 0.428 & 0.413 & 0.490 & 0.445 \\ 
    ~ & \textbf{720} & \textbf{MAE} & 0.429 & 0.437 & \underline{0.419} & \textbf{0.417} & 0.436 & 0.459 & 0.450 & 0.439 & 0.589 & 0.461 & 0.453 & 0.550 & 0.490 \\ 
    ~ & ~ & \textbf{MSE} & 0.463 & 0.490 & \underline{0.434} & \textbf{0.432} & - & 0.491 & 0.478 & 0.454 & 0.666 & 0.487 & 0.474 & 0.595 & 0.543 \\ 
    \midrule
    ~ & \textbf{96} & \textbf{MAE} & 0.270 & 0.282 & 0.269 & \underline{0.260} & 0.263 & 0.264 & 0.267 & \textbf{0.259} & 0.366 & 0.305 & 0.292 & 0.377 & 0.287 \\ 
    ~ & ~ & \textbf{MSE} & 0.200 & 0.205 & 0.195 & 0.189 & - & \underline{0.180} & 0.187 & \textbf{0.175} & 0.287 & 0.207 & 0.193 & 0.286 & 0.203 \\ 
    ~ & \textbf{192} & \textbf{MAE} & 0.315 & 0.318 & 0.303 & \textbf{0.300} & 0.309 & 0.309 & 0.309 & \underline{0.302} & 0.492 & 0.364 & 0.362 & 0.445 & 0.328 \\ 
    \textbf{ETTm2} & ~ & \textbf{MSE} & 0.269 & 0.261 & \underline{0.247} & \underline{0.247} & - & 0.250 & 0.249 & \textbf{0.241} & 0.414 & 0.290 & 0.284 & 0.399 & 0.269 \\ 
    ~ & \textbf{336} & \textbf{MAE} & \textbf{0.319} & 0.355 & \underline{0.333} & 0.334 & 0.349 & 0.348 & 0.351 & 0.343 & 0.542 & 0.422 & 0.427 & 0.591 & 0.366 \\ 
    ~ & ~ & \textbf{MSE} & \textbf{0.264} & 0.319 & \underline{0.291} & 0.295 & - & 0.311 & 0.321 & 0.305 & 0.597 & 0.377 & 0.369 & 0.637 & 0.325 \\ 
    ~ & \textbf{720} & \textbf{MAE} & \textbf{0.374} & 0.410 & \underline{0.377} & 0.386 & 0.415 & 0.407 & 0.403 & 0.400 & 1.042 & 0.524 & 0.522 & 0.735 & 0.415 \\ 
    ~ & ~ & \textbf{MSE} & \textbf{0.354} & 0.415 & \underline{0.355} & 0.372 & - & 0.412 & 0.408 & 0.402 & 1.730 & 0.558 & 0.554 & 0.960 & 0.421 \\ 
    \midrule
    ~ & \textbf{96} & \textbf{MAE} & \textbf{0.212} & 0.299 & 0.248 & 0.242 & - & \underline{0.240} & 0.272 & 0.285 & 0.314 & 0.329 & 0.282 & 0.345 & 0.308 \\ 
    ~ & ~ & \textbf{MSE} & \textbf{0.124} & 0.205 & 0.158 & 0.152 & - & \underline{0.148} & 0.168 & 0.195 & 0.219 & 0.237 & 0.197 & 0.247 & 0.193 \\ 
    ~ & \textbf{192} & \textbf{MAE} & \textbf{0.232} & 0.310 & 0.263 & 0.259 & - & \underline{0.253} & 0.289 & 0.289 & 0.322 & 0.330 & 0.285 & 0.355 & 0.315 \\ 
    \textbf{Electricity} & ~ & \textbf{MSE} & \textbf{0.138} & 0.220 & 0.174 & 0.171 & - & \underline{0.162} & 0.184 & 0.199 & 0.231 & 0.236 & 0.196 & 0.257 & 0.201 \\ 
    ~ & \textbf{336} & \textbf{MAE} & \textbf{0.249} & 0.323 & 0.278 & 0.278 & - & \underline{0.269} & 0.300 & 0.305 & 0.337 & 0.344 & 0.301 & 0.369 & 0.329 \\ 
    ~ & ~ & \textbf{MSE} & \textbf{0.155} & 0.236 & 0.191 & 0.192 & - & \underline{0.178} & 0.198 & 0.215 & 0.246 & 0.249 & 0.209 & 0.269 & 0.214 \\ 
    ~ & \textbf{720} & \textbf{MAE} & \textbf{0.291} & 0.347 & \underline{0.307} & 0.313 & - & 0.317 & 0.320 & 0.337 & 0.363 & 0.373 & 0.333 & 0.390 & 0.355 \\ 
    ~ & ~ & \textbf{MSE} & \textbf{0.211} & 0.270 & 0.229 & 0.236 & - & 0.225 & \underline{0.220} & 0.256 & 0.280 & 0.284 & 0.245 & 0.299 & 0.246 \\ 
    \midrule
    ~ & \textbf{96} & \textbf{MAE} & 0.223 & 0.212 & \textbf{0.203} & \underline{0.208} & - & 0.214 & 0.220 & 0.218 & 0.230 & 0.261 & 0.255 & 0.306 & 0.296 \\ 
    ~ & ~ & \textbf{MSE} & 0.180 & 0.173 & \underline{0.167} & 0.177 & - & 0.174 & 0.172 & 0.177 & \textbf{0.158} & 0.202 & 0.196 & 0.221 & 0.217 \\ 
    ~ & \textbf{192} & \textbf{MAE} & 0.267 & 0.250 & \textbf{0.241} & \underline{0.249} & - & 0.254 & 0.261 & 0.259 & 0.277 & 0.298 & 0.296 & 0.340 & 0.336 \\ 
    \textbf{Weather} & ~ & \textbf{MSE} & 0.235 & 0.216 & \underline{0.209} & 0.219 & - & 0.221 & 0.219 & 0.225 & \textbf{0.206} & 0.242 & 0.237 & 0.261 & 0.276 \\ 
    ~ & \textbf{336} & \textbf{MAE} & 0.291 & \underline{0.282} & \textbf{0.276} & 0.292 & - & 0.296 & 0.306 & 0.297 & 0.335 & 0.335 & 0.335 & 0.378 & 0.380 \\ 
    ~ & ~ & \textbf{MSE} & \textbf{0.252} & 0.260 & \underline{0.256} & 0.277 & - & 0.278 & 0.280 & 0.278 & 0.272 & 0.287 & 0.283 & 0.309 & 0.339 \\ 
    ~ & \textbf{720} & \textbf{MAE} & 0.356 & \textbf{0.322} & \underline{0.323} & 0.350 & - & 0.349 & 0.359 & 0.348 & 0.418 & 0.386 & 0.381 & 0.427 & 0.428 \\ 
    ~ & ~ & \textbf{MSE} & 0.356 & \textbf{0.320} & \underline{0.321} & 0.365 & - & 0.358 & 0.365 & 0.354 & 0.398 & 0.351 & 0.345 & 0.377 & 0.403 \\ 
\bottomrule
\end{tabular}%
}
\caption{Performance metrics for various models. Key: \textbf{Best results}, \underline{Second-best results.}}
  \label{tab:lsf_full_prediction_lengths}  
\end{table}

\end{document}